\documentclass[journal]{IEEEtran}

\ifCLASSINFOpdf
\else
\fi
\usepackage{cite}
\usepackage{amsmath,amssymb,amsfonts}
\usepackage{algpseudocode} 
\usepackage{algorithm} 
\usepackage{algorithmicx}
\usepackage{graphicx}
\usepackage{textcomp}
\usepackage{xcolor}
\usepackage{subfig}
\usepackage{tabularx,booktabs}
\usepackage{dblfloatfix}
\usepackage{soul}
\usepackage{xcolor} 
\usepackage{multirow}
\usepackage{caption}
\usepackage{threeparttable}
\usepackage{bbm}

\usepackage[switch]{lineno} 

\definecolor{mycolor}{RGB}{0, 27, 150} 
\definecolor{redcolor}{RGB}{200, 0, 0} 

\usepackage{xr}
\makeatletter

\newcommand*{\addFileDependency}[1]{
\typeout{(#1)}
\@addtofilelist{#1}
\IfFileExists{#1}{}{\typeout{No file #1.}}
}\makeatother

\newcommand*{\myexternaldocument}[1]{%
\externaldocument{#1}%
\addFileDependency{#1.tex}%
\addFileDependency{#1.aux}%
}

\myexternaldocument{supplementary}

\DeclareGraphicsExtensions{.png}

\begin{document}

%
\title{SSL-SoilNet: A Hybrid Transformer-based Framework with Self-Supervised Learning for Large-scale Soil Organic Carbon Prediction}
%
%
%

\author{Nafiseh~Kakhani,
        Moien~Rangzan,
        Ali~Jamali,
        Sara~Attarchi,
        Seyed~Kazem~Alavipanah, 
        Michael~Mommert,
        Nikolaos~Tziolas, and
        Thomas~Scholten


\thanks{N. Kakhani and T. Scholten are with the Department of Geosciences, Soil Science and Geomorphology, CRC 1070 RessourceCultures, and DFG Cluster of Excellence “Machine Learning”, University of Tübingen, 72070 Tübingen, Germany. (email: nafiseh.kakhani@uni-tuebingen.de) }

\thanks{M. Rangzan, S. Attarchi and S.K. Alavipanah are with the Department of Remote Sensing and GIS, Faculty of Geography, University of Tehran, Tehran, Iran.}

\thanks{A. Jamali is with the Department of Geography, Simon Fraser University, 8888 University Dr, Burnaby, BC V5A 1S6, Canada.}

\thanks{M. Mommert is with the Faculty of Geomatics, Computer Science and Mathematics, Stuttgart University of Applied Sciences, 70174 Stuttgart, Germany} 

\thanks{N. Tziolas is with the Southwest Florida Research and Education Center, Department of Soil, Water, and Ecosystem Sciences, Institute of Food and Agricultural Sciences, University of Florida, 2685 State Rd 29N, Immokalee, FL 34142, USA.}

\thanks{This research was funded by the Deutsche Forschungsgemeinschaft (DFG) [3150] for the project ‘MLTRANS-Transferability of Machine Learning Models in Digital Soil Mapping’ and a grant to Thomas Scholten (SCHO 739/21-1) and 'Machine Learning for Science'  which is part of Germany’s Excellence Strategy – EXC number 2064/1 – Project number 390727645. (\textit{Corresponding author: Nafiseh Kakhani})}
}

\newpage
\onecolumn 
\thispagestyle{empty} 

\vspace*{\fill} 

\begin{center}
    \footnotesize © 2024 IEEE. This is the author’s version of the work. It is posted here for your personal use. Not for redistribution. The definitive version will be published in IEEE Transactions on Geoscience and Remote Sensing (TGRS).\\[1em]
    This is the accepted version of the manuscript that has been accepted for publication in IEEE Transactions on Geoscience and Remote Sensing (TGRS). The final version, including the DOI, will be available through IEEE Xplore.
\end{center}

\vspace*{1cm} 

\newpage
\twocolumn
\setcounter{page}{1} 

\maketitle


\begin{abstract}
Soil Organic Carbon (SOC) constitutes a fundamental component of terrestrial ecosystem functionality, playing a pivotal role in nutrient cycling, hydrological balance, and erosion mitigation. Precise mapping of SOC distribution is imperative for the quantification of ecosystem services, notably carbon sequestration and soil fertility enhancement. Digital soil mapping (DSM) leverages statistical models and advanced technologies, including machine learning (ML), to accurately map soil properties, such as SOC, utilizing diverse data sources like satellite imagery, topography, remote sensing indices, and climate series. Within the domain of ML, self-supervised learning (SSL), which exploits unlabeled data, has gained prominence in recent years. This study introduces a novel approach that aims to learn the geographical link between multimodal features via self-supervised contrastive learning, employing pretrained Vision Transformers (ViT) for image inputs and Transformers for climate data, before fine-tuning the model with ground reference samples. The proposed approach has undergone rigorous testing on two distinct large-scale datasets, with results indicating its superiority over traditional supervised learning models, which depends solely on labeled data. Furthermore, through the utilization of various evaluation metrics (e.g., RMSE, MAE, CCC, etc.), the proposed model exhibits higher accuracy when compared to other conventional ML algorithms like random forest and gradient boosting. This model is a robust tool for predicting SOC and contributes to the advancement of DSM techniques, thereby facilitating land management and decision-making processes based on accurate information.

\end{abstract}


\IEEEpeerreviewmaketitle

\section{Introduction}

\IEEEPARstart{I}{n} recent years, deep learning (DL) have become an increasingly popular field of study. It has been used to model sequences and data with spatial context in computer vision, speech recognition and control systems, as well as in related scientific fields in physics, chemistry and biology \cite{reichstein2019deep}. One of the primary benefits of DL models, specifically convolutional neural networks (CNNs), is their ability to learn abstract hierarchical representations of the data, which enables networks to uncover spatial, spectral, and temporal patterns hidden in the data. This enables researchers and engineers to streamline the information extraction processing chain, potentially integrating multi-modal data fusion, feature extraction, and inference duties into a single, holistic end-to-end learning framework \cite{persello2022deep}. The convergence of DL approaches for both spatial learning and sequence learning has generated significant interest due to its remarkable resemblance to many dynamic geoscience challenges (e.g. \cite{yuan2020deep}).

\begin{figure*}[t]
\centering
\includegraphics[width=6in]{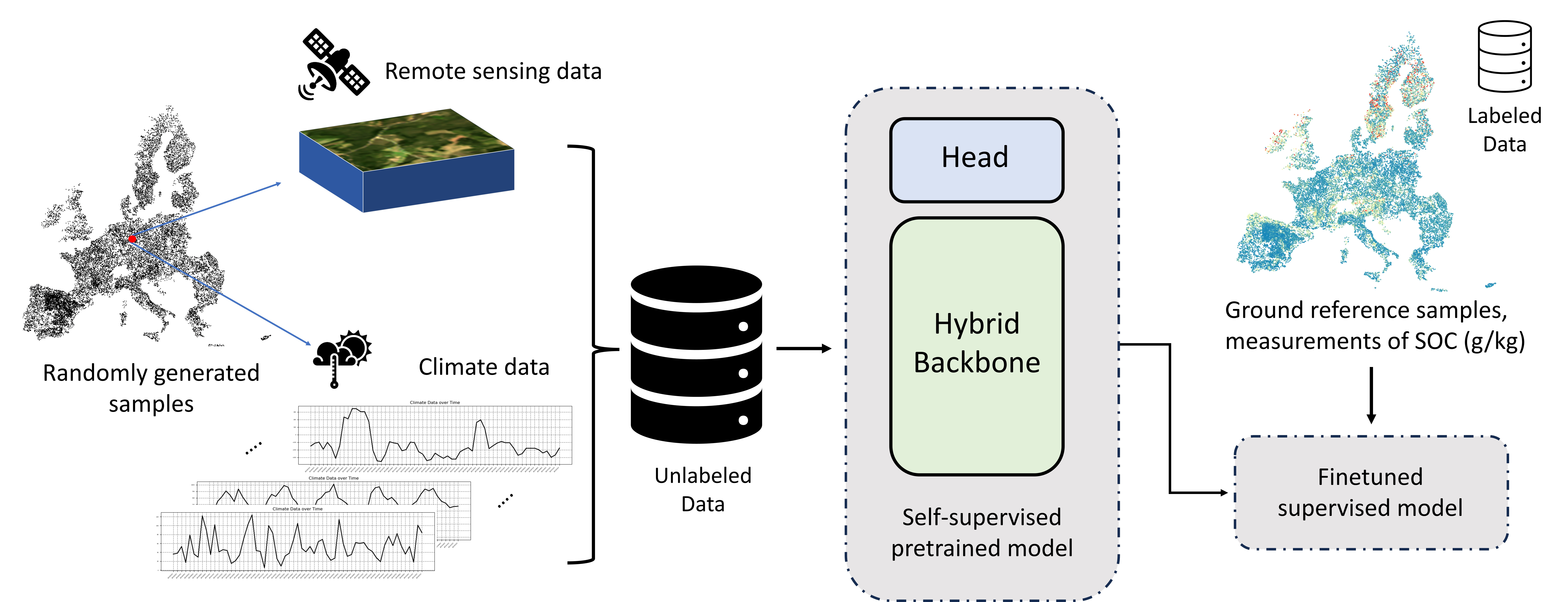}
\caption{Conceptual framework of this study. A large number of unlabeled data is leveraged for self-supervised training, followed by fine-tuning on a limited number of available soil samples serving as ground truth.}
\label{fig:SoilNet1}
\end{figure*}

There are a wide range of applications in geoscience that can be addressed via DL among which we can mention anomaly detection, dynamic modelling, semantic segmentation, hyperspectral data analysis and digital soil mapping (DSM) \cite{taghizadeh2021enhancing}. DSM is one of the most prevalent methods for mapping soil properties, and its use continues to increase \cite{arrouays2014globalsoilmap}. It refers to the process of producing high-resolution digital maps of soil types and properties using statistical and ML techniques to integrate multiple forms of data that largely control soil forming processes, such as topography, vegetation, climate, and parent material \cite {mcbratney2003digital}. These maps can be used for a variety of purposes, including precision agriculture, land-use planning, environmental management, and natural disaster risk assessment. As soil is essential for ecosystems and plays a significant role in carbon cycling, food security, biodiversity, and human activity, DSM enables a more efficient and accurate characterization of soil variability and provides a more comprehensive understanding of soil distribution and properties over vast areas. In particular, DSM has proven to be a valuable tool for mapping soil organic carbon (SOC) content (e.g. \cite{taghizadeh2020improving, yang2022effectiveness, ladoni2010estimating }). SOC is a crucial component of the global carbon cycle and plays a vital role in regulating climate change. Two-thirds of the carbon in the terrestrial environment is found in the form of SOC \cite{post1982soil}. 

\hl{}SOC prediction using DSM is an example of a regression problem which can be effectively addressed by an ML approach as proven by previous studies like \cite{grimm2008soil, hengl2017soilgrids250m}. With the advent of remote sensing satellites and the development of this technology over the past decade, methods based on ML have become even more widespread in the field of DSM using remote sensing images or other products such as vegetation indices, at local and global scales (e.g. \cite{zepp2023optimized, dasgupta2023developing, lemercier2022multiscale}). 

Yet, classical ML faces some criticisms due to its limitations.
One significant limitation is its failure to consider the dynamic nature of certain input data, notably climate data utilized extensively in DSM, which ought to be treated as time series data. However, classical ML treats each observation as a single value, disregarding the temporal relationships between observations. Due to many inherent factors, such as abnormal climatic conditions, the accuracy of monotemporal data is basically limited \cite{yang2023digital}. To overcome these limitations, advanced deep learning techniques, such as transformers \cite{vaswani2017attention}, provide a powerful solution. Transformers, originally developed for natural language processing tasks, have shown exceptional performance in time series analysis due to their ability to capture long-term dependencies within data \cite{cachay2022climformer}. Transformers leverage self-attention mechanisms weigh the importance of different points in the time series, here climate data, allowing it to focus on relevant data points regardless of their position in the sequence \cite{prapas2023televit}.

\begin{figure*}[t]
\centering
\includegraphics[width=7.3in]{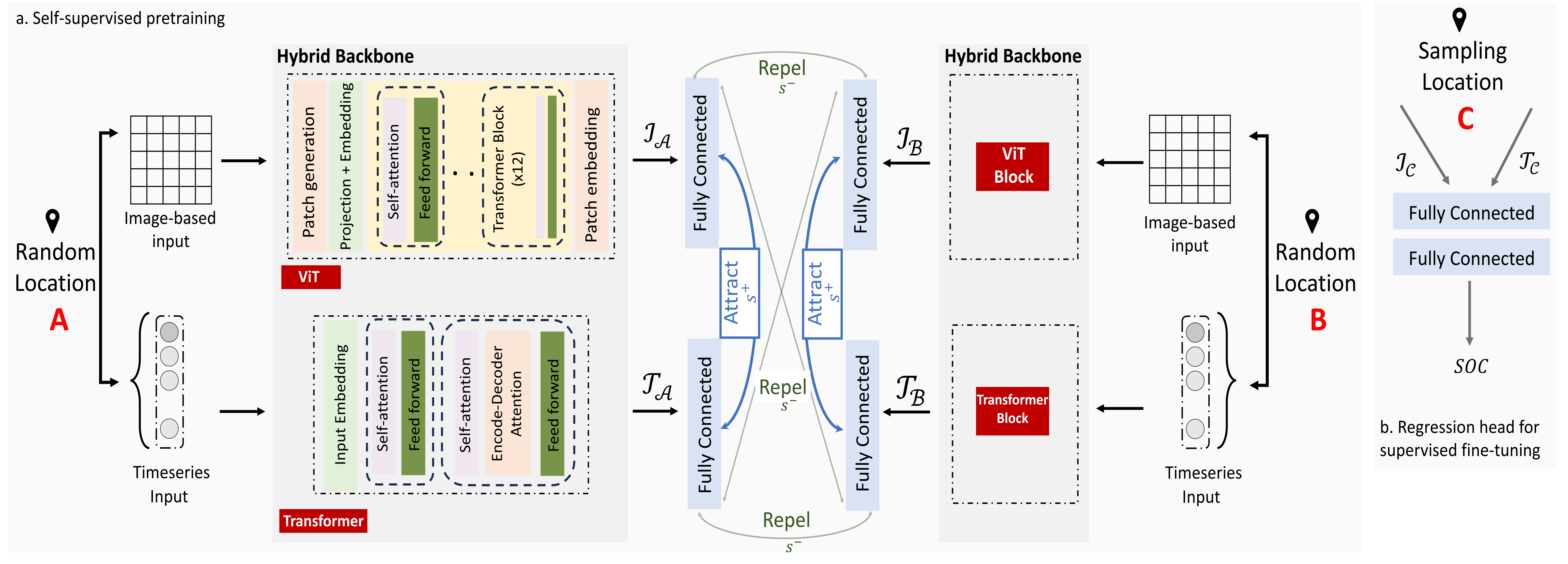}
\caption{Simplified architecture illustrating the proposed approach. The training process involves two steps. In the initial phase, contrastive learning is utilized, wherein both a ViT and a transformer are trained. They process pairs of image-based and time-series input data obtained through random sampling. The ViT and transformer extract intermediate representations, $\mathcal{I}$ and $\mathcal{T}$, used for self-supervised training. In the second step, fine-tuning is performed using the same architecture with ground truth samples. A regression head is added to predict the target output, SOC.} 
\label{fig:SSLsection}
\end{figure*}

Classical ML also faces another drawback as it overlooks spatial and neighborhood details, often referred to as conceptual or geometric information. This is a basic limitation, as many remote sensing features exhibit spatial correlation, meaning that their values at one location are often related to the values at nearby locations and this information can be highly beneficial for prediction \cite{behrens2018spatial}. In this case, spatial autocorrelation in the residuals of spatial environmental models could be accounted for by using features from multiple scales \cite{behrens2019relevant} and a variogram can be used to determine an effective scale. 
Several studies in the DSM field have attempted to address this issue using DL models. Specifically in \cite{taghizadeh2020multi}, a CNN-based architecture with shared layers was utilized to predict particle size fractions of clay, sand, and silt across six standard layers. More recently, Tziolas et al. introduced a dual input DL architecture that simultaneously combined Sentinel-2 data and environmental covariates through two distinct neural network branches to estimate SOC \cite{TZIOLAS2024soc}. Nowadays, with the emergence of vision transformers, Jagetia et al. applied this method to soil classification, demonstrating superior performance compared to other models \cite{jagetia2022visual}. Additionally, in a study by Jin et al., Vis-NIR spectra were leveraged in conjunction with the Swin Transformer \cite{liu2021swin} to predict soil properties \cite{jin2023innovative}.

The efficacy of transformers rely on the availability of a large number of data samples for training a robust model \cite{xu2024self}. However, acquiring soil property information is often expensive and time-consuming, involving in-situ surveys and extensive laboratory analyses. Consequently, ground truth samples may be insufficient in certain regions, posing challenges in training reliable DL models. To mitigate this dependence on high-quality datasets, self-supervised learning emerges as a promising alternative. In self-supervised learning, a model is initially pre-trained on unlabeled data, leveraging inherent data structures or patterns to learn valuable representations or features \cite{chen2020big}. Subsequently, the pre-trained model undergoes fine-tuning using labeled data samples to adapt to specific downstream tasks, such as image classification.

Given the often easy accessibility of unlabeled data, such as RS imagery and climate data, self-supervised pre-training proves effective in enabling the model to acquire reliable and adaptable feature representations \cite{scheibenreif2022self}. This approach is particularly advantageous when the downstream task lacks sufficient labeled data, as pre-trained models can prevent overfitting and yield robust performance post fine-tuning. In self-supervised learning, contrastive learning stands out as a widely employed technique \cite{chen2020simple}. Contrastive learning aims to learn representations shared among similar samples (positive pairs) while differentiating them from dissimilar samples (negative pairs) \cite{scheibenreif2022self}. Typically, positive pairs consist of two random "views" of the same sample through random data augmentation, while negative pairs arise from other samples in the dataset. The learning objective in contrastive learning is to minimize a contrastive loss, promoting the closeness of representations for positive pairs and pushing apart representations for negative pairs. However, these methods cannot be directly applied to multimodal data, as image-specific augmentation techniques are not suitable for temporal sequences. The application of contrastive learning to address soil property estimation remains a relatively unexplored area.

Thus, in this study, we present a novel hybrid network designed for SOC prediction on a large scale, leveraging a substantial number of unlabeled samples during the pre-training phase, without requiring extensive laboratory analysis. This network called SSL-SoilNet, utilizing a self-supervised contrastive learning approach. The methodology comprises two main phases. Initially, self-supervised contrastive learning is applied to a large unlabeled samples that includes both remote sensing and climate data. This phase focuses on leveraging the inherent structure and patterns within the multimodal data for unsupervised feature learning. Subsequently, the pre-trained self-supervised model is transitioned into the supervised segment, where ground reference samples are incorporated for fine-tuning. The second phase focuses on refining the model for SOC prediction by using labeled data.

This framework proficiently extracts spatial and contextual information from remote sensing images while capturing time-series information from climate data. Our hybrid approach uniquely tackles the challenge of creating meaningful connections between image and climate data. In contrast to preceding models, our framework demonstrates an understanding of the geographical linkage between climate and image data for each sample. This is achieved through the identification of similarities between image and climate data (positive and negative pairs), enabling the effective utilization of the advantages inherent in the multimodal dataset employed in this study. Consequently, the main contributions of this article are mentioned as follows:

\begin{enumerate}

  \item A pioneering hybrid multimodal model has been developed to predict SOC by integrating both RS and climate data.
  
  \item A novel self-supervised contrastive learning framework is introduced, leveraging a substantial volume of unlabeled data during the initial phase and subsequently fine-tuning with a limited set of ground reference samples.
  
  \item The research diverges from conventional contrastive learning methodologies by employing an inventive approach that calculates similarities between climate and image data at specific locations. This departure from traditional image augmentation techniques represents a noteworthy innovation in the field.
  
\end{enumerate}

\section{Proposed Approach}
\subsection{An Overview of SSL-SoilNet Architecture}

The SSL-SoilNet architecture, illustrated in Figure \ref{fig:SoilNet1}, consists of two main phases: a self-supervised pretraining using contrastive learning with a substantial number of unlabeled, randomly selected samples, followed by fine-tuned supervised learning using ground reference samples. At each location, we gathered two primary types of features: image-based remote sensing $(I)$ and time-series climate data $(T)$. For each feature type, a unique  backbone was utilized.  

\subsection{Self-Supervised Contrastive Learning}
\label{sec:ContrastiveSSL}

Self-supervised contrastive learning is a paradigm that entails the unsupervised pre-training of models using unlabeled data. This approach is characterized by the formulation of a task wherein the model endeavors to discriminate between positive pairs (representing similar instances, $s^+$) and negative pairs (representing dissimilar instances, $s^-$) within the given unlabeled data. The central idea, derived from the theoretical mathematical framework outlined in \cite{sohn2016improved}, involves optimizing the model through a contrastive loss function, which quantifies the divergence or convergence between the representations of input pairs. The general formulation of the contrastive loss $\ell$ is expressed as:

\begin{equation}
\ell_{i, j}= -\log \frac{\exp \left(\operatorname{sim}\left(s_i, s_j\right) / \tau\right)}{\sum_{k=1}^{2 N} \mathbbm{1}_{[k \neq i]} \exp \left(\operatorname{sim}\left(s_i, s_k\right) / \tau\right)} \label{eq:simclr_loss}
\end{equation}

\noindent where;\\
$~~~~~~{s}_i$ and ${s}_j$ represent a positive pair, $s^+$\\
$~~~~~~{s}_k$ are negative (contrastive) pairs, $s^-$\\
$~~~~~~\mathbbm{1}$ is the indicator function\\
$~~~~~~\tau$ is temperature parameter

The temperature $\tau$ allows us to balance the influence of many dissimilar representations versus a similar one. This idea was introduced first in \cite{chen2020simple} as SimCLR. The similarity function, $\operatorname{sim}(\cdot, \cdot)$, is cosine similarity as defined by:

\begin{equation}
\operatorname{sim}\left(u, v\right)=\frac{u^{\top} \cdot v}{\left\|u\right\| \cdot\left\|v\right\|} \label{eq:sim}
\end{equation}

\begin{algorithm}[b]
\caption{SSL-SoilNet's training algorithm}\label{alg:SSL}
\begin{algorithmic}[0]

\State \textbf{Input:} Image-based features $I$ and time-series features $T$ for each sample $\textbf{x}$, batch size $N$, temperature parameter $\tau$, Transformer $(\textbf{Trans})$ and ViT $(\textbf{ViT})$ networks, and a multilayer perceptron $(\textbf{MLP})$. 
\State \textbf{Output:} Updated $\textbf{Trans}$ and $\textbf{ViT}$ networks.

\For{each sample in minibatch $\{x_k = [I_k, T_k]\}^{N}_{k=1}$}
    \For{all $k \in \{1, \ldots, N\}$}
        \State $\mathcal{I}_{k} = \textbf{ViT}(I_{k})$ \Comment{\textcolor{gray}{Image-based inputs}}
        \State $s_{2k-1} = \textbf{MLP}(\mathcal{I}_{k})$ 

        \State $\mathcal{T}_{k} = \textbf{Trans}(T_{k})$ \Comment{\textcolor{gray}{Time-series inputs}}
        \State $s_{2k} = \textbf{MLP}(\mathcal{T}_{k})$ 
    \EndFor
     
    \For{all $i \in \{1, \ldots, 2N\}$ and $j \in \{1, \ldots, 2N\}$}
        \State Compute cosine similarity between $s_{i}$ and $s_{j}$ using Eq.\ref{eq:sim}.
    \EndFor
     
    \State Define $\ell_{i,j}$ (Eq.\ref{eq:simclr_loss})
    \State Minimize $\mathcal{L} = \frac{1}{2N} \sum_{(i, j) \text{ positive pair}} \left(\ell_{i, j} + \ell_{j, i}\right)$ to update $\textbf{Trans}$ and $\textbf{ViT}$. 
\EndFor
\end{algorithmic}
\end{algorithm}

As depicted in Figure \ref{fig:SSLsection}, a pivotal departure from the classical SimCLR methodology is observed concerning positive pairs denoted as $s^+$. In contrast to the conventional SimCLR approach where positive pairs involve augmented images, here, these pairs comprise representations derived from image-based inputs, $(I)$, and temporal sequences, $(T)$, originating from the identical spatial location. This distinction holds significance as it introduces a novel dimension to the concept of positive pairs within the SimCLR framework. Rather than relying solely on image augmentations to diversify positive pairs, the inclusion of temporal information alongside spatial representation enhances the model's capacity to encapsulate both static and dynamic aspects inherent to the given spatial context. Consequently, the positive pairs, now comprising not only visual variations but also temporal dynamics, contribute to a more comprehensive and contextually rich feature extraction process. This adaptation of positive pairs aligns with the input data's intrinsic nature, emphasizing the capture of spatiotemporal relationships in SimCLR. It represents a deliberate refinement catering to diverse data modalities, especially in soil properties prediction scenarios where both spatial and temporal cues collectively shape the underlying information structure. The pseudocode outlining the training procedure is detailed in Algorithm \ref{alg:SSL}.

\begin{table}[t]
    \renewcommand{\arraystretch}{1.3}
    \caption{Statistical summaries of the two ground reference samples of SOC utilized in this study}
    \label{tab: GT_descp}
    \centering

    \begin{threeparttable}
        \begin{tabular}{cccccccc}
            \hline
            & Mean & s.d. & Min. & Q1 & Median & Q3 & Max.\\
            \hline
            LUCAS$^1$ & 29.86 & 24.43 & 0.10 & 12.5 & 20.40 & 38.60 & 87.00 \\
            RaCA$^2$ & 182.69 & 416.61 & 0.17 & 52.76 & 86.54 & 149.69 & 4115 \\
            \hline
        \end{tabular}
        \footnotesize
        \begin{tablenotes}
            \item $^1$The SOC content is measured in (g/kg). $^2$The SOC stock is expressed in (Mg/ha).
        \end{tablenotes}
    \end{threeparttable}
\end{table}

\begin{figure}[!t]
\centering
\includegraphics[width=3.5in]{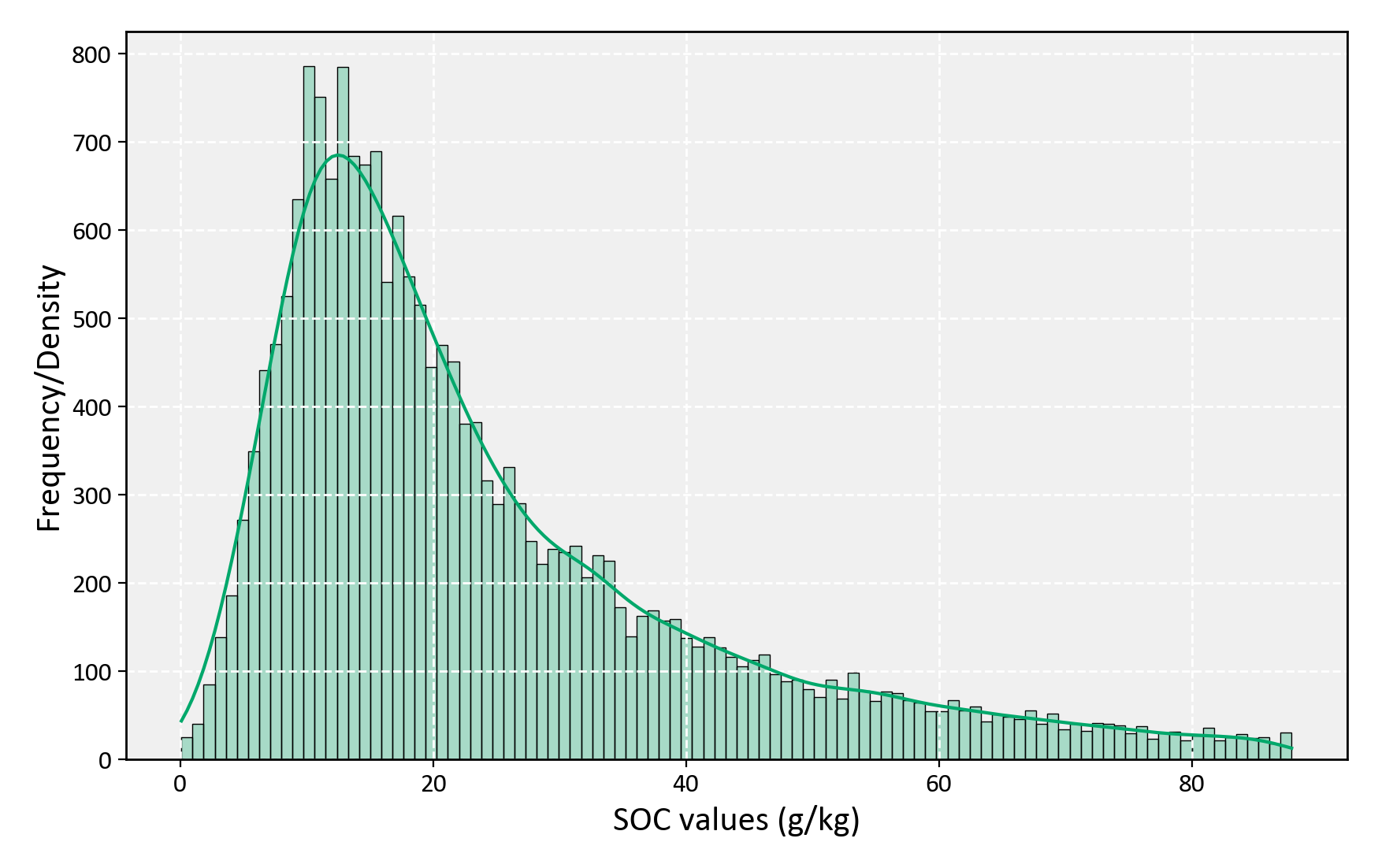}
\caption{Histogram and kernel density estimation plot depict the distribution of SOC (g/kg) values for LUCAS dataset.}
\label{fig:LUCAS_dist}
\end{figure}

\begin{figure}[!t]
\centering
\includegraphics[width=3.5in]{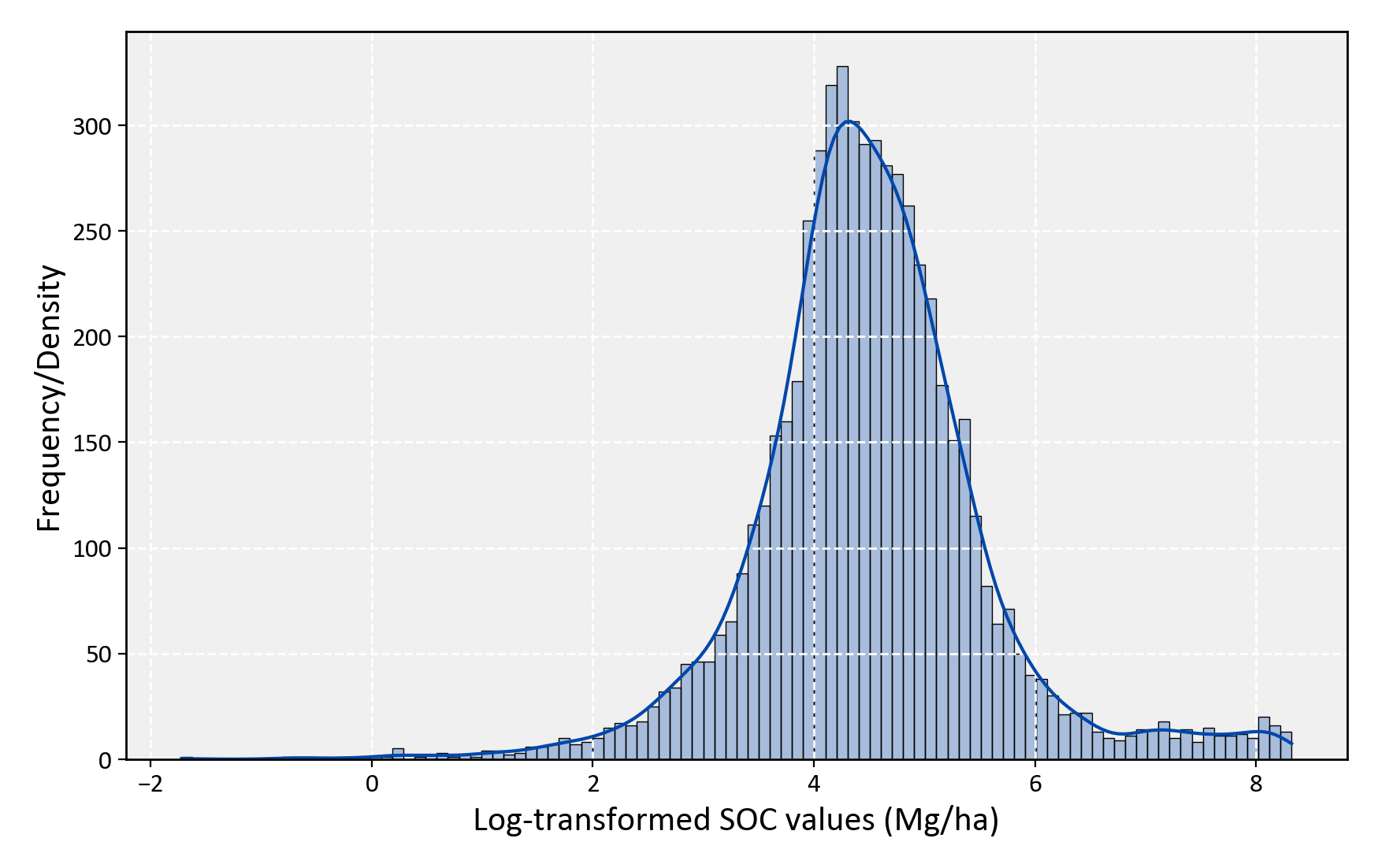}
\caption{Histogram and kernel density estimation plot depict the distribution of Log-transformed SOC (Mg/ha) values for RaCA dataset.}
\label{fig:RaCA_dist}
\end{figure}

\subsection{Fine-tuned Supervised Model}

The self-supervised model is now poised for fine-tuning to perform a particular task, such as regression. In the supervised fine-tuning phase, the objective is to regress the joined feature vectors to SOC values. For this purpose, the intermediate representations for both image-based and timeseries inputs, denoted as $\mathcal{I}$ and $\mathcal{T}$, as seen in Figure \ref{fig:SSLsection}, are fed into a 2-layer MLP block, which is trained using labeled data. Given the skewed distribution observed in soil samples, the Root Mean Squared Logarithmic Error (RMSLE) is selected as the loss function. RMSLE has the potential to mitigate the effects of this skewed distribution and can be defined as:

\begin{equation}
RMSLE = \sqrt{\frac{1}{n}\sum_{i=1}^{n}(\log(y_i + 1) - \log(\hat{y}_i + 1))^2} \label{eq:rmsle}
\end{equation}

\noindent where; \\
$~~~~~~~~\hat{y}_i$ is the predicted and $y_i$ is observed value.\\
$~~~~~~~~\log(.)$ denotes the natural logarithm function.\\

The $+1$ inside the logarithm function is added to avoid taking the logarithm of zero or negative values, which would be undefined.

\section{Data Description and Prepossessing}

\begin{figure}[!t]
\centering
\includegraphics[height=5in]{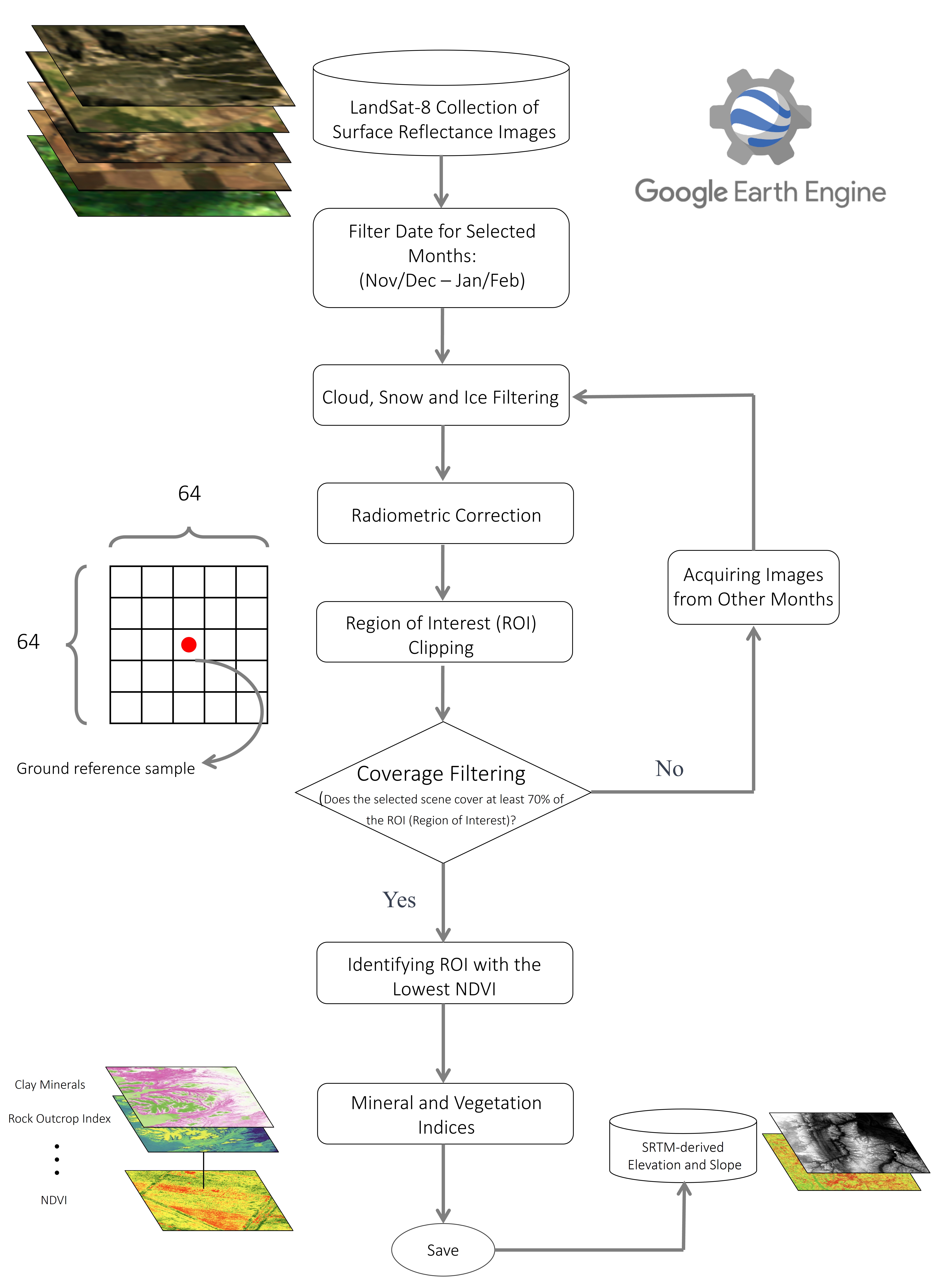}
\caption{The process of raster-based data preparation.}
\label{fig: data_preparation1}
\end{figure}

This section provides an overview of ground reference samples and the input features used in SSL-SoilNet, which are classified into four categories. The first three categories consist of image raster-based inputs for the image-based backbone, including Landsat-8 bands, mineral and vegetation indices, and topography. The diagram illustrating the preparation of these inputs is shown in Figure \ref{fig: data_preparation1}. The fourth category comprises climate features, which serve as inputs for the transformer component. 

\subsection{Ground Reference Samples}
\subsubsection{LUCAS Dataset}

The LUCAS programme was established in 2001 under the supervision of Eurostat, the statistical office of the European Union. This program aimed to conduct an area frame survey, focusing on visually evaluating factors relevant to agricultural policies. Starting from 2006, the sampling methodology adopted a regular grid system spanning the entire territory of the EU, with each grid cell measuring 22 km. Eurostat collaborated with the European Commission's Directorates-General for the Environment and the Joint Research Centre to develop a specific component within the LUCAS survey known as 'LUCAS-Topsoil.' This component focused on assessing the topsoil, specifically within the 0 to 20 cm depth range, to generate harmonized and comparable data on soil across Europe, supporting evidence-based policy making \cite{orgiazzi2018lucas}. For our analysis, we utilized the LUCAS dataset from 2015, which includes information on SOC for all EU countries and has around 21000 samples. In line with the findings of a previous study predicting SOC in Germany \cite{Sakhaee2022}, which highlighted the improved accuracy of separate modeling for mineral and organic carbon compared to a combined model, we have specifically focused on samples representing mineral soil, identified by values below 87 (g/kg).

\subsubsection{RaCA Dataset}

The Rapid Carbon Assessment (RaCA) project, initiated by the Soil Science Division of the Natural Resources Conservation Service (NRCS), aims to systematically document SOC throughout the conterminous United States (CONUS) within a specific timeframe. Our study draws upon around 6200 soil field observations of SOC stock (Mg/ha) aggregated to a depth of 100 cm through fixed-depth sampling. These observations were strategically collected across diverse land cover types, soil classifications, and ecological regions. For comprehensive insights into the sampling and analytical methodologies employed in RaCA, interested readers are referred to the detailed exposition provided by \cite{west2013rapid}. The statistical information of the data are summarised in table \ref{tab: GT_descp}. The histogram and kernel density functions of LUCAS and RaCA datasets are also provided in Figures \ref{fig:LUCAS_dist} and \ref{fig:RaCA_dist}.

\begin{figure}[!t]
\centering
\includegraphics[width=3.5in]{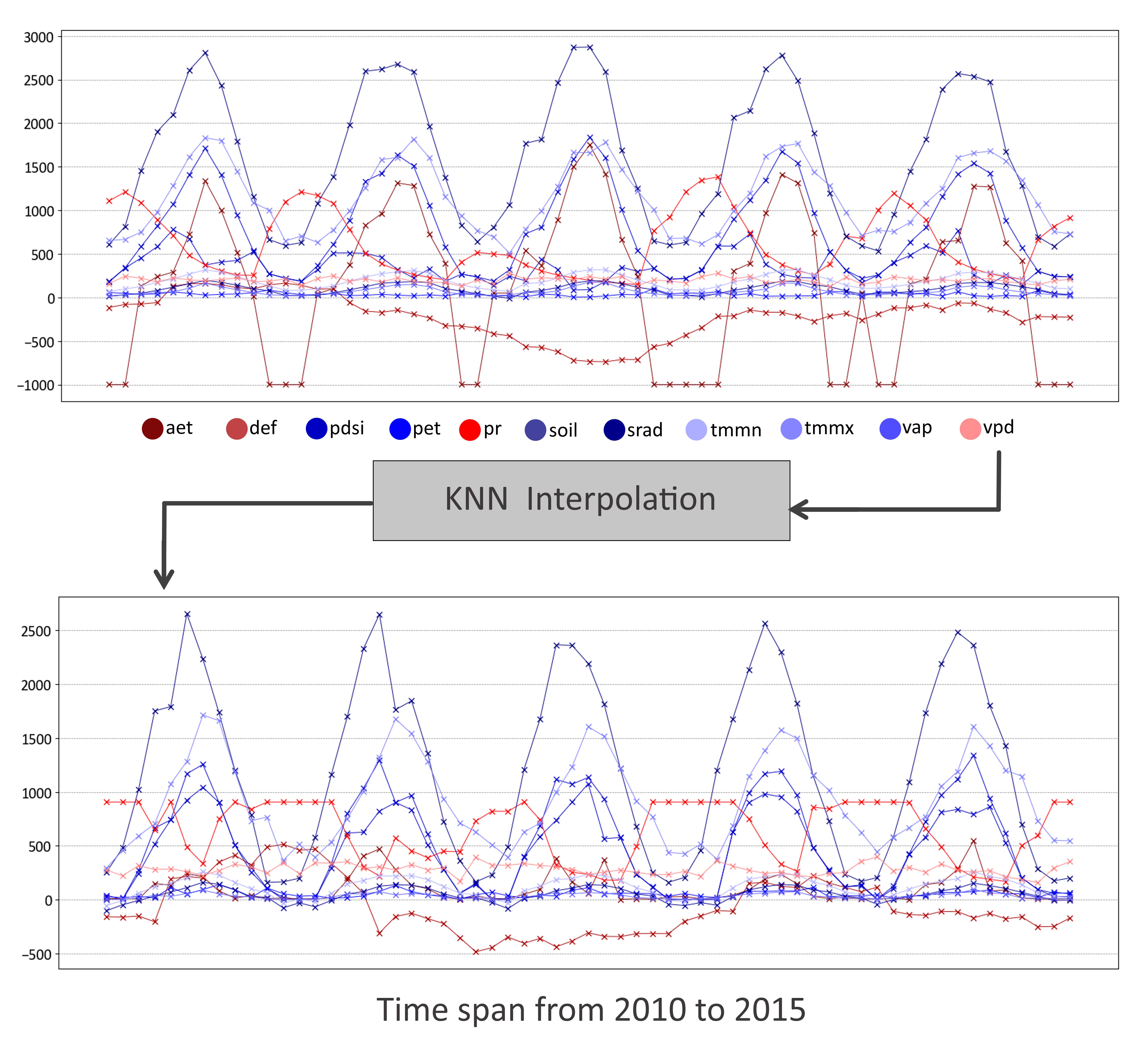}
\caption{Timeseries (Climate) data preparation.}
\label{fig:data_preparation2}
\end{figure}

\subsection{Input Features}
\subsubsection{Landsat-8 Bands}

For this study, Landsat-8 surface reflectance images from the year 2015 were utilized due to their proximity to the sampling period for both datasets. To mitigate the influence of vegetation on our analysis, we focused on acquiring images captured during the cold months (November/ December and January/February), which are less likely to be affected by dense vegetation cover. Additionally, we applied a filtering process to exclude images with significant cloud cover or snow, ensuring that they accounted for less than 10 percent of the total image area. During the cold months in most European countries, there is a high chance of not finding any image due to cloud cover. In such cases, we selected images from other months of the year with the lowest NDVI value. Our overall process is also consistent with the finding of the previous study in \cite{dvorakova2023improving}. To obtain the necessary data for each sample point, we downloaded Landsat image patches centered around the sample location, employing a window size of 64 by 64 pixels. This carefully designed data pipeline enabled us to obtain suitable soil-focused images for our analysis, minimizing confounding factors and maximizing the relevance of the obtained data.

\subsubsection{Mineral and Vegetation Indices}

In our study, we incorporated a set of indices that encompass both soil and vegetation characteristics. These indices, namely clay minerals \cite{alasta2011using}, Ferrous Minerals \cite{segal1982theoretical}, Carbonate Index, Rock Outcrop Index, and normalized difference vegetation index (NDVI), provide valuable insights into the soil and vegetation properties of the study area. The names and equations for each index have been provided in Table \ref{tab:Cov_descr}. These indices offer valuable insights into various soil and vegetation properties, allowing us to better explore and understand the relationship between the study area's soil composition.

\begin{table*}
\caption{ Input feature description}
\label{tab:Cov_descr}
\renewcommand{\arraystretch}{1.2} 
\resizebox{\textwidth}{!}{%
\begin{tabularx}{\textwidth}{l l l l l l}
\toprule
\textbf{No} & \textbf{Category} & \textbf{Feature} & \textbf{Description} & \textbf{Unit} \\
\midrule
1 & \multirow{12}{*}{\rotatebox[origin=c]{90}{\centering L8 images + indices}} & B1 & Ultra Blue & \multirow{7}{*}{\rotatebox[origin=c]{90}{\centering \({{\text{W}}} / {{\text{m}^2 \cdot \text{sr} \cdot \mu\text{m}}} \)}} 
\\
2 && B2 & Blue && 
\\
3 && B3 & Green && 
\\
4 && B4 & Red && \\
5 && B5 & NIR && \\
6 && B6 & SWIR1 && \\
7 && B7 & SWIR2 && \\
8 && Clay Minerals & (SWIR1 - SWIR2)/(SWIR1 + SWIR2)  & \multirow{5}{*}{\rotatebox[origin=c]{90}{\centering Unitless}} \\
9 && Ferrous Minerals & (NIR - SWIR1)/(NIR + SWIR1) &   \\
10 && Carbonate Index & (R-G)/(R+G) & \\
11 && Rock Outcrop Index & (SWIR1-G)/(SWIR+G) &   \\
12 && NDVI & (NIR-R)/(NIR+R) &   \\
\midrule
13 & \multirow{2}{*}{\rotatebox[origin=c]{90}{\centering Topo }} & Elevation & Elevation & m   \\
14 && Slope & Slope & Percent  \\
\midrule
15 & \multirow{5}{*}{\rotatebox[origin=c]{90}{\centering PrimClim.}}& Minimum temperature & Minimum temperature (tmmn) & °C   \\
16 && Maximum temperature & Maximum temperature (tmmx) & °C  \\
17 && Vapor pressure & Vapor pressure (vpd) & kPa  \\
18 && Precipitation accumulation & Precipitation accumulation (pr) & mm  \\
19 && Surface radiation & Downward surface shortwave radiation (srad) & W/m\textasciicircum{}2  \\
\midrule
20 & \multirow{5}{*}{\rotatebox[origin=c]{90}{\centering SecClim.}} & Actual evapotranspiration & Actual evapotranspiration, derived using a one-dimensional soil water balance model (aet) & mm   \\
21 && pdsi & Palmer Drought Severity Index (pdsi) & Unitless  \\
22 && Climate water deficit & Climate water deficit, derived using a one-dimensional soil water balance model (def) & mm \\
23 && Reference evapotranspiration & Reference evapotranspiration (pet) & mm  \\
24 && Vapor pressure deficit & Vapor pressure deficit (vap) & kPa   \\
25 && Soil moisture & Soil moisture (soil) & mm \\
\hline
\end{tabularx}%
}
\end{table*}

\subsubsection{Topography}

We utilized digital elevation data acquired from the Shuttle Radar Topography Mission (SRTM) for our analysis. This dataset was the result of a global research endeavor that generated digital elevation models with nearly global coverage. We obtained the SRTM-V3 product (SRTM Plus) provided by NASA JPL, which offers a resolution of 1 arc-second, equivalent to approximately 30 meters. In addition to the elevation data, we incorporated slope too. These variables play a significant role in shaping the distribution of soil across the landscape by influencing processes such as overland flow and erosion, which, in turn, impact the dynamics of SOC \cite{Carter1991, Pei2010, Scholten2017}. 
By integrating these specific topographic factors, we hope to achieve a thorough understanding of the complex relationship between terrain characteristics and the dynamics of SOC in the designated study region.

\subsubsection{Climate Features}
\label{sec:climate}

In order to investigate the diverse environmental conditions that either promote or impede climate regulation, we utilized a comprehensive set of parameters provided by TerraClimate \cite{Abatzoglou2018}. These parameters were derived from gridded meteorological data, utilizing a climatically aided spatiotemporal interpolation technique applied to the WorldClim datasets \cite{hijmans2005very}. This approach facilitated the estimation of monthly time series. The selection of environmental factors for this study included crucial variables categorized into two groups: 1) primary climate variables, comprising maximum temperature, minimum temperature, vapor pressure, precipitation accumulation and downward surface shortwave radiation and 2) derived or secondary variables, encompassing reference evapotranspiration, actual evapotranspiration, climate water deficit, soil moisture, Palmer drought severity index (pdsi), and vapor pressure deficit. The rationale behind including these specific variables stems from their widely acknowledged influence on SOC dynamics \cite{Fick2017, Sakhaee2022} as well as the underlying processes that contribute to the ecosystem's climate regulation function \cite{Tamburini2020, Yang2020}. To obtain the necessary data, we downloaded meteorological data from 2010 to 2015. Any missing values present in the dataset were interpolated using k-nearest neighbour (KNN) to ensure the completeness of the temporal records (See Figure. \ref{fig:data_preparation2}). A comprehensive list of all input features is provided in Table \ref{tab:Cov_descr}.

\section{Experiments and analysis}
\subsection{Implementaion Detials}

For each configuration, the network was trained using a 5-fold cross-validation approach, and the optimal outcomes are documented for test samples. We divided the data into train (60\%), validation (20\%), and test (20\%) sets. The optimization method employed was Adam \cite{kingma2014adam}. The convergence of all applied methods required 50 epochs. The weights were initialized using a random initialization, and the initial learning rate was set to $1 \times 10^{-4}$. The learning rate was gradually reduced over time to enhance stability and refine the model's performance. All experimental procedures were conducted utilizing the PyTorch deep learning library, on a single NVIDIA RTX 3090 with 24 GB GDDR6X memory. Each run for the LUCAS dataset takes 70 minutes, and for the RaCA dataset, it takes 59 minutes to complete.

\subsection{Evaluation Metrics}

When evaluating the performance of predictive models, several commonly used evaluation metrics include Mean Absolute Error (MAE), the coefficient of determination (\(R^2\)), Root Mean Square Error (RMSE), Ratio of Performance to Interquartile distance (RPIQ) and Concordance Correlation Coefficient (CCC).
MAE is calculated by taking the average of the absolute differences between the predicted values and the true values:

\begin{equation}
MAE = \frac{1}{n}\sum_{i=1}^{n}|y_i - \hat{y}_i| 
\label{MAE}
\end{equation}

\(R^2\) (also known as the coefficient of determination) is a statistical measure that represents the proportion of the variance in the dependent variable that can be explained by the independent variables in a regression model. It is commonly used to assess the goodness of fit of a regression model. This value is calculated using the following equation:

\begin{equation}
R^2 = 1 - \frac{{\sum_{i=1}^{n}(y_i - \hat{y}_i)^2}}{{\sum_{i=1}^{n}(y_i - \bar{y})^2}}
\end{equation}

where:
\(n\) is the number of observations, \(y_i\) is the observed value of the dependent variable for observation \(i\), \(\hat{y}_i\) is the predicted value of the dependent variable for observation \(i\) based on the regression model, and \(\bar{y}\) is the mean of the observed values of the dependent variable.

RMSE is calculated by taking the square root of the average of the squared differences between the predicted values and the true values:

\begin{equation}
RMSE = \sqrt{\frac{1}{n}\sum_{i=1}^{n}(y_i - \hat{y}_i)^2} \label{RMSE}
\end{equation}

RPIQ represents the spread of the population and is calculated using the following equation \cite{bellon2010critical}:

\begin{equation}
\label{RPIQ}
\text{RPIQ} = \frac{Q_3 - Q_1}{\text{RMSE}}
\end{equation}

The values \(Q_1\) and \(Q_3\) represent the 25th and 75th percentiles of the true samples, respectively, defining the interquartile distance.

CCC is a measure of the agreement between the predicted values and the true values. It takes into account both the mean difference and the variance difference between the predicted and true values:

\begin{equation}
CCC = \frac{2\rho\sigma_y\sigma_{\hat{y}}}{\sigma_y^2 + \sigma_{\hat{y}}^2 + (\mu_y - \mu_{\hat{y}})^2} \label{CCC}
\end{equation}
where \(\rho\) represents the correlation coefficient between the predicted and observed values, \(\sigma_y\) and \(\sigma_{\hat{y}}\) are the standard deviations of the observed and predicted values respectively, and \(\mu_y\) and \(\mu_{\hat{y}}\) are the means of the observed and predicted values respectively.

These metrics provide quantitative measures to assess the accuracy, correction, agreement, and calibration of predictive models compared to the observed values.

\subsection{Results}

The outcomes of the conducted experiments are comprehensively consolidated in Tables \ref{tab:Results_LUCAS} and \ref{tab:RaCA_results}. The details of each experiment group will be discussed in the following sections.


\begin{table*}[t]
\caption{Analysis of performance across different network configurations for supervised and self-supervised learning; an ablation study using the LUCAS$^{*}$ dataset.} \label{tab:Results_LUCAS}
\centering
\renewcommand{\arraystretch}{1} 
\begin{tabular*}{\linewidth}{@{\extracolsep{\fill}}ccccccccccc}
\toprule
\multicolumn{8}{c}{\textbf{Input Features}} & \multicolumn{1}{c}{ \textbf{ Evaluation Metrics} } \\
\cmidrule(lr){3-6}
\cmidrule(lr){7-11}
Approach & Config & L8+RS ind  & Topo & PrimClim. & SecClim. & MAE $\downarrow$ & $\mathrm{R}^2(\%)$$\uparrow$ & RMSE $\downarrow$ & RPIQ $\uparrow$  & CCC $\uparrow$ \\
\midrule
\multirow{5}{*}{\rotatebox[origin=c]{90}{\centering Supervised}} & \multirow{5}{*}{\centering ViT + Trans.} & 1 & 1 & 1 & 1 & 13.70 & 39 & 19.04 & 1.48  & 0.55  \\
& & 1 & 0 & 1 & 1 & 13.95 & 37 & 19.28 & 1.44  & 0.54\\
& & 1 & 1 & 0 & 1 & 15.16 & 27 & 20.89 & 1.24  & 0.42\\
& & 1 & 1 & 1 & 0 & 15.15 & 27 & 20.90 & 1.21  & 0.41\\
& & 1 & 1 & 0 & 0 & 15.99 & 19 & 21.88 & 1.09  & 0.32\\
\cmidrule(lr){1-6}
\cmidrule(lr){7-11}
\multirow{1}{*}{\rotatebox[origin=c]{0}{\centering SSL}} & \multirow{1}{*}{\centering ViT + Trans.} & 1 & 1 & 1 & 1 & \textbf{12.82} & \textbf{43} & \textbf{18.31} & 1.40  & \textbf{0.60} \\
\midrule
\midrule
\multirow{5}{*}{\rotatebox[origin=c]{90}{\centering Supervised}} & \multirow{5}{*}{\centering ViT + LSTM} & 1 & 1 & 1 & 1 & 14.35 & 34 & 19.85 & 1.34  & 0.50  \\
& & 1 & 0 & 1 & 1 & 14.67 & 30 & 20.36 & 1.20  & 0.46\\
& & 1 & 1 & 0 & 1 & 14.32 & 33 & 19.97 & 1.22  & 0.48\\
& & 1 & 1 & 1 & 0 & 14.90 & 28 & 20.66 & 1.18  & 0.43\\
& & 1 & 1 & 0 & 0 & 15.11 & 27 & 20.77 & 1.28  & 0.32\\
\cmidrule(lr){1-6}
\cmidrule(lr){7-11}
\multirow{1}{*}{\rotatebox[origin=c]{0}{\centering SSL}} & \multirow{1}{*}{\centering ViT + LSTM} & 1 & 1 & 1 & 1 & 13.58 & 38 & 19.16 & 1.35  & 0.56 \\
\midrule
\midrule
\multirow{5}{*}{\rotatebox[origin=c]{90}{\centering Supervised}} & \multirow{5}{*}{\centering CNN + Trans.} & 1 & 1 & 1 & 1 & 14.72 & 32 & 20.11 & 1.32  & 0.48  \\
& & 1 & 0 & 1 & 1 & 15.82 & 30 & 20.34 & 1.28  & 0.43  \\
& & 1 & 1 & 0 & 1 & 14.17 & 38 & 19.13 & 1.61  & 0.58 \\
& & 1 & 1 & 1 & 0 & 14.78 & 38 & 19.23 & \textbf{1.63}  & 0.54\\
& & 1 & 1 & 0 & 0 & 14.58 & 31 & 20.30 & 1.14  & 0.46 \\
\cmidrule(lr){1-6}
\cmidrule(lr){7-11}
\multirow{1}{*}{\rotatebox[origin=c]{0}{\centering SSL}} & \multirow{1}{*}{\centering CNN + Trans.} & 1 & 1 & 1 & 1 & 13.40 & 40 & 18.92 & 1.39  & 058 \\
\midrule
\midrule
\end{tabular*}
\footnotesize
    \begin{tablenotes}
           \item $^*$The unit for MAE, RMSE and CCC is (g/kg). RPIQ is unitless.
    \end{tablenotes}
\end{table*}
\begin{table*}[!t]
\caption{Analysis of performance across different network configurations for supervised and self-supervised learning; an ablation study using the RaCA$^{*}$ dataset.}
\label{tab:RaCA_results}
\centering
\renewcommand{\arraystretch}{1} 
\begin{tabular*}{\linewidth}{@{\extracolsep{\fill}}ccccccccccc}
\toprule
\multicolumn{8}{c}{\textbf{Input Features}} & \multicolumn{1}{c}{ \textbf{ Evaluation Metrics} } \\
\cmidrule(lr){3-6}
\cmidrule(lr){7-11}
Approach & Config & L8+RS ind  & Topo & PrimClim. & SecClim. & MAE $\downarrow$ & $\mathrm{R}^2 (\%)$$\uparrow$ & RMSE $\downarrow$ & RPIQ $\uparrow$  & CCC $\uparrow$ \\
\midrule

\multirow{5}{*}{\rotatebox[origin=c]{90}{\centering Supervised}} & \multirow{5}{*}{\centering ViT + Trans.} & 1 & 1 & 1 & 1 & 133.98 & 17 & 360.04 & 0.34  & 0.27  \\
& & 1 & 0 & 1 & 1 & 130.54 & 17 & 360.58 & 0.31  & 0.27  \\
& & 1 & 1 & 0 & 1 & 130.61 & 16 & 363.12 & 0.28  & 0.25\\
& & 1 & 1 & 1 & 0 & 133.35 & 15 & 364.66 & 0.31  & 0.24\\
& & 1 & 1 & 0 & 0 & 137.08 & 9 & 377.99 & 0.29  & 0.12 \\

\cmidrule(lr){1-6}
\cmidrule(lr){7-11}
\multirow{1}{*}{\rotatebox[origin=c]{0}{\centering SSL}} & \multirow{1}{*}{\centering ViT + Trans.} & 1 & 1 & 1 & 1 & \textbf{125.59} & \textbf{21} & \textbf{351.66} & 0.29  & \textbf{0.32} \\
\midrule
\midrule
\multirow{5}{*}{\rotatebox[origin=c]{90}{\centering Supervised}} & \multirow{5}{*}{\centering ViT + LSTM} & 1 & 1 & 1 & 1 & 128.81 & 16 & 363.14 & 0.28 & 0.26  \\
& & 1 & 0 & 1 & 1 & 128.93 & 16 & 363.23 & 0.28  & 0.26  \\
& & 1 & 1 & 0 & 1 & 127.46 & 15 & 363.83 & 0.27  & 0.25\\
& & 1 & 1 & 1 & 0 & 129.67 & 15 & 364.54 & 0.29  & 0.24\\
& & 1 & 1 & 0 & 0 & 137.68 & 6 & 383.50 & 0.27  & 0.09 \\
\cmidrule(lr){1-6}
\cmidrule(lr){7-11}
\multirow{1}{*}{\rotatebox[origin=c]{0}{\centering SSL}} & \multirow{1}{*}{\centering ViT + LSTM} & 1 & 1 & 1 & 1 & 126.85 & 17 & 361.55 & 0.31 & 0.32 \\
\midrule
\midrule
\multirow{5}{*}{\rotatebox[origin=c]{90}{\centering Supervised}} & \multirow{5}{*}{\centering CNN + Trans.} & 1 & 1 & 1 & 1 & 133.04 & 16 & 362.84 & \textbf{0.37} & 0.25  \\
& & 1 & 0 & 1 & 1 & 134.08 & 14 & 365.99 & 0.24  & 0.26  \\
& & 1 & 1 & 0 & 1 & 130.06 & 15 & 364.34 & 0.21  & 0.26\\
& & 1 & 1 & 1 & 0 & 128.65 & 14 & 365.84 & 0.25  & 0.24\\
& & 1 & 1 & 0 & 0 & 132.05 & 8 & 378.81 & 0.21  & 0.15 \\
\cmidrule(lr){1-6}
\cmidrule(lr){7-11}
\multirow{1}{*}{\rotatebox[origin=c]{0}{\centering SSL}} & \multirow{1}{*}{\centering CNN + Trans.} & 1 & 1 & 1 & 1 & 126.79 & 19 & 354.54 & 0.32 & 0.29 \\
\midrule
\midrule
\end{tabular*}
\footnotesize
    \begin{tablenotes}
           \item $^*$The unit for MAE, RMSE and CCC is (Mg/ha). RPIQ is unitless.
    \end{tablenotes}
\end{table*}

\subsubsection{Baselines}
\label{sec:baselines}

Our study includes three primary network configurations for both the LUCAS dataset (see Table \ref{tab:Results_LUCAS}) and the RaCA dataset (see Table \ref{tab:RaCA_results}), denoted as ViT+Trans., ViT+LSTM, and CNN+Trans. These configurations represent distinct hybrid architectural approaches we adopted for our analysis. Specifically, we utilized the ViT and a conventional CNN, here ResNet-101, to process image data. For the analysis of time-series data, we substitute long short-term memory (LSTM) networks over Transformers due to LSTM's established efficacy in managing sequential information \cite{hochreiter1997long}. Irrespective of the selected method (supervised or SSL) and the type of input features, our findings generally indicate that the ViT+Trans. configuration outperforms the other two configurations across most evaluated metrics.

\subsubsection{Contrastive SSL vs. Supervised Learning}

Each network configuration introduced in the previous section \ref{sec:baselines}, was evaluated using two distinct methodologies. Initially, we conducted supervised training using only the ground reference samples available for each dataset. Subsequently, we integrated unlabeled samples into a self-supervised contrastive learning model, as described in Section \ref{sec:ContrastiveSSL}. 

For the LUCAS dataset, approximately 70,000 unlabeled randomly selected samples were utilized, whereas for the RaCA dataset, around 30,000 were employed to prepare our pre-trained model. To guarantee relevance to soil, we evaluated the land cover classes within each downloaded image patch. If the majority of a Landsat-8 image patch was identified as irrelevant land-cover classes such as built-up areas or permanent water bodies, that sample was excluded from our analysis. We have used using ESA land cover product \cite{esa_worldcover} for this purpose.

In both datasets, the performance of SSL-SoilNet was superior to that of supervised-SoilNet, demonstrating the pre-trained model's effectiveness in linking image-based and climate-based information geographically. This integration of information contributed to enhanced prediction accuracy. A notable observation was made with the RaCA dataset results, where the improvement in evaluation metrics was substantial. Specifically, the RMSE decreased by 8.38 Mg/ha, the $R^2$ increased by 4\%, and the MAE was reduced by 8.39 Mg/ha. This marked enhancement may be attributed to the model's ability to address severe data imbalance, especially in the upper quantiles of SOC ranges. Essentially, leveraging information from random samples allowed the model to better understand patterns in samples with high SOC levels. It is important to highlight that the RPIQ metric does not show improvement, which is attributed to its definition focusing on the population spread rather than solely on accuracy. A model exhibiting a low RMSE can still result in a high RPIQ, suggesting that RMSE and RPIQ metrics should be analyzed in conjunction.

\subsubsection{Ablation Study}

To evaluate the contribution of input features in our study, we conducted an ablation study by categorizing our input features into four groups: Landsat images and their derived indices, topographic information (i.e., elevation and slope), and two types of climate information (i.e., primary and secondary variables) as introduced in Section \ref{sec:climate}. We tested each feature group across different configurations, including ViT+Transformer, ViT+LSTM, and CNN+Transformer models in Tables \ref{tab:Results_LUCAS} and \ref{tab:RaCA_results}. As expected, the exclusion of all climate information resulted in the lowest performance across all configurations, underscoring the significance of using timeseries information. This is because soil properties are influenced not only by surface information from satellite images but also by climatic conditions. Incorporating secondary climate variables yields marginally improved outcomes in terms of RMSE, CCC, and \(R^2\) compared to primary climate variables, indicating a contributory role of this feature group. Topographic information consistently enhanced prediction accuracy for both datasets. The omission of primary and secondary climate variables sometimes led to a reduction in the MAE, for example, in the supervised training of ViT+LSTM for the LUCAS dataset and supervised training of ViT+Trans. for the RaCA dataset. This is possibly due to the complex nature of time series data, which may contain noisy information. Removing such noisy data can, in some instances, improve overall prediction accuracy.

\begin{figure*}[t]
\centering
\includegraphics[width=7in]{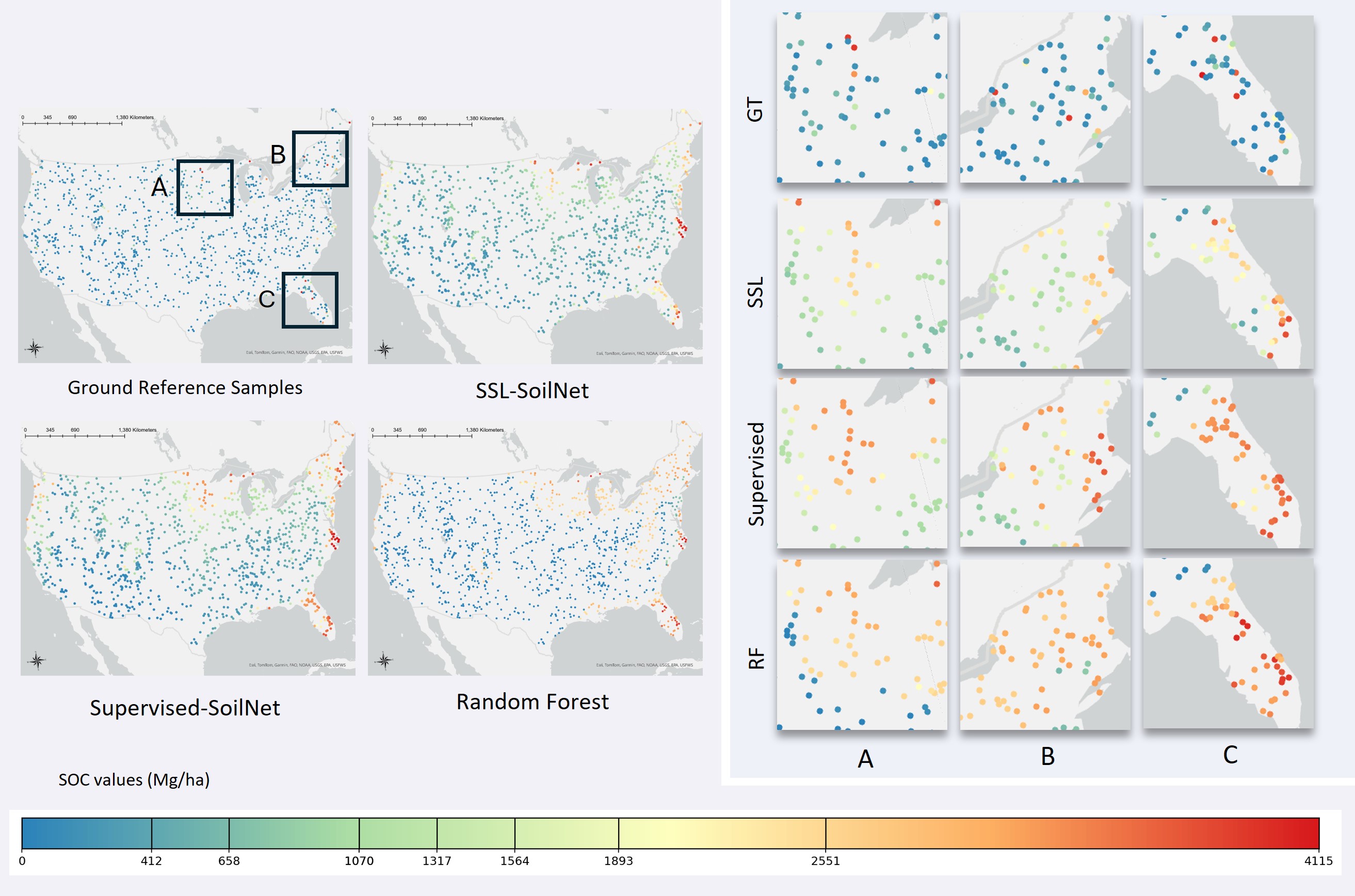}
\caption{The distribution of SOC (Mg/ha) measurements for test samples and the results from implemented models, SSL and supervised SoilNet, and RF, on the left side. Areas A, B, and C highlight regions with data heterogeneity. On the right side, zoomed-in sections compare the performance of different models against the ground truth across four rows, with SSL-SoilNet's performance showing close alignment with the ground reference samples.}
\label{fig:fig_map_RaCA}
\end{figure*}

\begin{figure*}[h]
\centering
\includegraphics[height=5in]{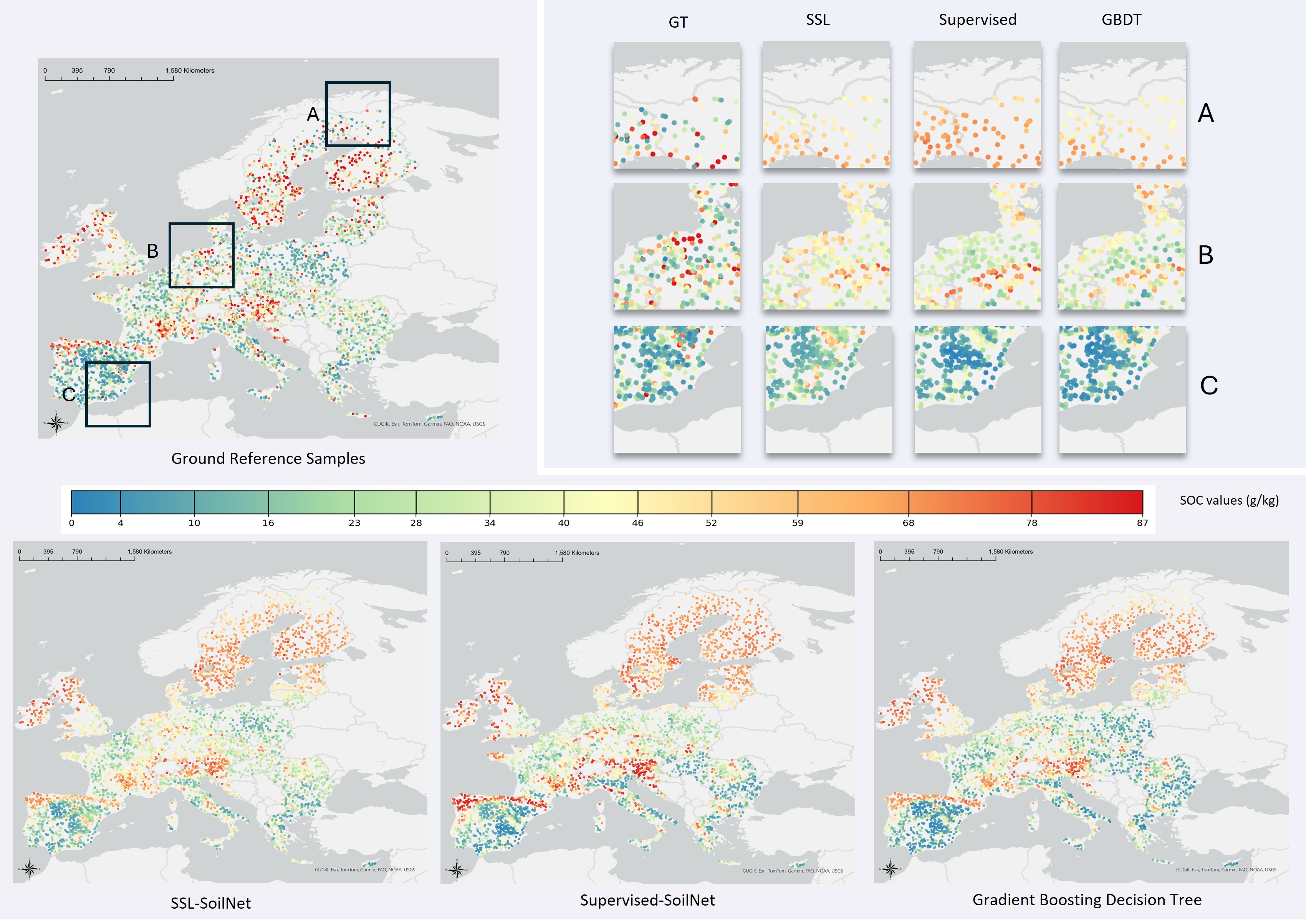}
\caption{The distribution of SOC content measurements (g/kg) for the test samples is displayed at the top left. The results from the implemented models—SSL and supervised SoilNet, along with Gradient Boosting Decision Tree (GBDT)—are located in the lower part. Areas A, B, and C highlight regions with data heterogeneity. On the right side, zoomed-in sections compare the performance of different models against the ground truth across three rows. SSL-SoilNet's performance shows close alignment with the ground reference samples.}
\label{fig:fig_map_LUCAS}
\end{figure*}

\subsection{Comparison to the other ML Models}

To comprehensively assess our model's performance, we adopted a straightforward approach as the first step. We assumed the most basic scenarios for prediction models: for the LUCAS dataset, which exhibits a distribution close to normal, we used the mean as the simplest prediction for all available samples. Conversely, for the RaCA dataset, which demonstrates severe skewness, we employed the median of the available ground truth. Subsequently, we computed all evaluation metrics. 
Although the naive model is not expected to achieve high performance, as evidenced by the zero value for \( R^2 \), it offers valuable insights into the efficacy of our proposed approach and other machine learning models when benchmarked against the most basic scenarios.

In the second step, we compared our results with those from seven commonly used ML models. The first model we evaluated was Random Forest (RF), a popular ML algorithm often applied to regression tasks, especially in the field of soil science (e.g, \cite{szatmari2021estimating}). The second model, known as Deep Forest, is an advanced version of RF. It is a decision tree ensemble method that requires fewer hyperparameters than deep learning models \cite{zhou2019deep}. Additionally, we explored two variations of gradient boosting: XGBoost \cite{chen2016xgboost} and Gradient Boosting Decision Tree (GBDT) \cite{friedman2002stochastic}. XGBoost is an optimized distributed gradient boosting library known for enhancing efficiency and performance. GBDT is also a powerful technique that improves prediction accuracy by building decision trees sequentially, focusing on minimizing errors from previous trees. We also evaluated Extremely Randomized Trees (ERT) \cite{geurts2006extremely}, a variation of random forest that introduces additional randomness into the model, potentially increasing the diversity among the trees. The other ML models used for comparison were Support Vector Regressor (SVR) and Cubist, which is a rule-based model that combines decision trees and linear regression \cite{emadi2020predicting}. The hyperparameters for all implemented models were determined using randomized grid search.

Tables \ref{tab: LUCAS_ML} and \ref{tab: RaCA_ML} present the analysis of two datasets, LUCAS and RaCA, providing insightful observations on the performance of various models. Across both datasets, SSL-SoilNet, the self-supervised contrastive learning model, achieved the highest accuracy as measured by RMSE and \(R^2\). This highlights the superiority of our proposed approach, largely due to the incorporation of additional information from unlabeled data during training. Additionally, for the LUCAS dataset, SSL-SoilNet surpassed other models in MAE, CCC, and RPIQ metrics, indicating that our approach excels in accuracy, provides more realistic prediction intervals (as calculated through RPIQ), and ensures greater concordance between predictions and observations (measured by CCC), all while maintaining the lowest mean average error. Supervised SoilNet also showed commendable performance, surpassing other ML models. It's important to note that the \( R^2 \) values are low in certain cases, such as SVR and ERT, because these models struggled to capture the nonlinearity inherent in the data, which arises from the complexity of the soil system \cite{li2022soil}.

\begin{table}[!t]
\renewcommand{\arraystretch}{1.3}

\caption{Performance results for SSL and supervised SoilNet, and other common ML methods on the LUCAS dataset.}
\label{tab: LUCAS_ML}
\centering

\begin{tabular}{ c c c c c c c c}
\hline
\textbf{Model} & {MAE}$\downarrow$ & {\(R^2\)(\%)}$\uparrow$ & {RMSE}$\downarrow$ & {RPIQ}$\uparrow$  & {CCC}$\uparrow$ \\
\hline
SSL & \textbf{12.82} & \textbf{43} & \textbf{18.31} & \textbf{1.48} & \textbf{0.60} \\
Supervised & 13.70 & 39 & 19.04 & 1.40 &  0.55  \\
RF & 15.64 & 28 & 20.64 & 1.30 &  0.42\\
DeepRF & 13.63 & 32 & 20.12 & 0.92 & 0.47 \\
XGB & 14.53 & 28 & 20.66 & 1.35 & 0.56\\
GBDT & 14.30 & 38 & 19.14 & 1.44 &  0.52 \\
ETR & 15.83 & 27 & 20.77 & 1.29 &  0.39\\
SVR & 14.56 & 22 & 21.48 & 0.88 & 0.37\\
Cubist & 13.30 & 33 & 20.01 & 1.19 & 0.58\\
Basic & 19.21 & 0 & 24.43 & 0.7 & $~$0 \\
\hline
\end{tabular}
\end{table}

The scenario for the RaCA dataset was somewhat different, marked by significant positive skewness in value distribution, as detailed in Table \ref{tab: GT_descp}. Here, the RF model recorded the highest RPIQ, highlighting its capacity to generate wide prediction intervals, even though its predictions were less precise compared to those from SSL and Supervised SoilNet. DeepForest, an improved version of RF, achieved the best MAE, while XGBoost secured the highest CCC. These findings illustrate the varied strengths of different ML methodologies in SOC prediction, suggesting no single approach dominates across all evaluation metrics. Despite this, DL models, especially SSL-SoilNet, have proven to offer the most accurate predictions.

The pronounced skewness of the RaCA dataset has also led to significant variability in performance metrics among all implemented methods. For example, the disparity between the best and worst RMSE scores (noted between SSL-SoilNet and XGBoost, respectively) was 47.26 Mg/ha, with a $\sim$20 \% difference in \(R^2\) scores, highlighting the impact of skewed distributions on the consistency of model performance across various evaluation metrics. Furthermore, the SVR and Cubist models, exhibiting the lowest \(R^2\) value among the other ML models, indicated a deviation from the trend observed in the RaCA samples.

\begin{table}[!t]
\renewcommand{\arraystretch}{1.3}

\caption{Performance results for SSL and supervised SoilNet, and other common ML methods on the RaCA dataset.}
\label{tab: RaCA_ML}
\centering

\begin{tabular}{ c c c c c c c c}
\hline
\textbf{Model} & {MAE}$\downarrow$ & {\(R^2(\%)\)}$\uparrow$ & {RMSE}$\downarrow$ & {RPIQ}$\uparrow$  & {CCC}$\uparrow$ \\
\hline
SSL & 125.59 & \textbf{21} & \textbf{351.66} & 0.29 & 0.32 \\
Supervised & 130.54 & 17 & 360.58 & 0.31 &  0.27  \\
RF & 158.16 & 10 & 374.24 & \textbf{0.74} &  0.16\\
DeepRF & \textbf{116.98} & 9 & 376.05 & 0.22 & 0.16 \\
XGB & 167.74 & 1 & 398.92 & 0.30 &  \textbf{0.40}\\
GBDT & 155.60 & 7 & 380.60 & 0.33 &  0.40 \\
ETR & 156.85 & 9 & 377.94 & 0.51 &  0.13\\
SVR & 121.76 & -0.03 & 402.81 & 0.12 & 0.02\\
Cubist & 132.67 & 0.04 & 387.67 & 0.21 &0.22\\
Basic & 169.42 & 0 & 417.38 & 0.3 &  $~$0\\
\hline
\end{tabular}
\end{table}

\subsection{Discussion}

We have illustrated the spatial distribution of observed SOC for test samples, alongside the predicted SOC from both SSL and Supervised SoilNet, as well as the top-performing ML methods, RF for RaCA and GBDT for LUCAS datasets, across our study areas, which encompass the CONUS and Europe in Figures \ref{fig:fig_map_RaCA} and \ref{fig:fig_map_LUCAS}. To enhance the visualization of the various model performances, we plotted the predicted values against the observed values in Supplementary Figures \ref{fig:S1_LUCAS_reg} and \ref{fig:S2_RaCA_reg}. The map of residuals are also included in Supplementary Figure \ref{fig:error}. 

In our analysis of the RaCA dataset, regions such as the northeastern area, west coastal area, and rocky mountain area exhibit notable variations in SOC values \cite{cao2019spatial}. Specifically regarding mountainous regions, this variability can be partly ascribed to the limited availability of SOC observations at greater depths. The acquisition of samples from these deeper layers is resource-intensive and presents significant challenges \cite{wang2024upscaling}. Additionally, Mountain soils within the CONUS are characterized by their tendency to be shallow, thin, and coarse-textured. Consequently, variations in these properties are anticipated to have a more significant influence on SOC dynamics in these areas \cite{egli2016soils}. A notable discovery from our study is the discrepancy between the output of RF and DL models. This indicates that RF failed to capture the variability within these areas accurately.

We focused on three distinct areas characterized by varying climatic conditions and land cover classes to assess the performance of our implemented methodologies. Area A, located in the midwestern part of the United States, features land cover classes predominantly consisting of wetlands and tree covers \cite{USGS_LCMAP_2017}. This region's land cover is indicative of its ecological diversity and complexity. Area B, situated in the the northeastern United States, is characterized by wetlands and forests, along with relatively high precipitation and low average temperatures \cite{ClimateGov_Maps_2020}. Conversely, Area C is distinguished by its higher temperatures and above-average precipitation levels, with land cover primarily comprising wetlands, croplands, and developed areas in the southeastern United States. 

Despite the inherent differences among these areas, a commonality emerges in their heterogeneity concerning SOC values. Each area contains samples with both exceptionally low and high SOC values, presenting challenges for any ML/DL models in achieving accurate predictions. This variability underscores the complexity of SOC prediction and the necessity for sophisticated models capable of accommodating such diverse environmental conditions. The outcomes obtained from the RF model serve as evidence supporting this claim, as the final map generated by RF failed to accurately capture the data trends across Areas A, B, and C. This inadequacy highlights the model's limitation in reflecting the intricate spatial variations inherent to each area. In contrast, Supervised-SoilNet demonstrated a marginally improved performance, particularly for Area A, as depicted on the right side of Figure \ref{fig:fig_map_RaCA}. Supervised-SoilNet's slight improvement was not enough to accurately represent the correct data trends on a local scale. However, SSL-SoilNet outperformed the other models in accurately capturing the data trends across all designated areas. SSL-SoilNet demonstrates superior performance in dealing with the spatial complexity and variability of SOC values, offering a more precise and reliable prediction capability.

In the context of the LUCAS dataset, all models demonstrated the capability to capture data trends at a continental scale. In Europe, mountains and specifically the European Alps are of great significance. Forest soils within this region have the highest organic carbon concentrations \cite{baritz2010carbon}, where low air temperatures slow down the decomposition of organic material, leading to significant accumulation of soil organic matter. Additionally, grassland soils are considered one of the most specific biological hotspots, abundant in organic carbon \cite{garcia2024rapid}. Due to the limited accessibility and extensive spatial heterogeneity of these regions \cite{prietzel2014organic}, they could be among the most challenging areas.
Utilizing a considerable quantity of unlabeled samples during the self-supervised training in SSL-SoilNet provides the model with an opportunity to acquire insights into these regions even in the absence of plentiful ground truth samples. This facilitates more accurate predictions.

Detailed examinations were conducted for three distinct areas within Europe. Area A is situated in the northern part of Europe, specifically Finland, Norway and Sweden, characterized predominantly by wetlands, high precipitation, and low temperatures. Area B, located in Germany and the Netherlands, is primarily composed of grassland, cropland, and wetlands, coupled with below-average temperatures and relatively high precipitation. Conversely, Area C in Spain experiences low precipitation and high temperatures, with its land cover classes including cropland, grassland, and bareland.

For Area A, none of the implemented models were fully successful in accurately reflecting the trends observed in the ground reference samples. This difficulty in prediction for Area A aligns with previous studies that have identified it as a region with high uncertainty in SOC predictions \cite{kakhani2024uncertainty}. However, the maps generated by SSL-SoilNet showed a closer resemblance to the reference samples than those produced by both supervised training and the GBDT model, indicating a relatively better performance. A similar observation was made for Area B, where SSL-SoilNet outperformed other models in matching the reference data trends. In the case of Area C, which presents a distinct environmental profile from Areas A and B, the only model that demonstrated proficiency in identifying samples with high SOC values was SSL-SoilNet. This finding underscores the effectiveness of SSL-SoilNet in dealing with the complex variability of SOC across different landscapes, particularly in areas where environmental conditions significantly differ.

An intriguing finding from both datasets is the association between challenging areas and the presence of land cover class wetlands. Within these regions, soils have been influenced by decades of cold and anaerobic conditions, resulting in significant heterogeneity in composition \cite{baird2009understanding}. This diversity, reflecting a range of soil microbial activities and decelerated organic matter decomposition rates \cite{barreto2022decomposition}, presents challenges in accurately assessing SOC and predicting its distribution in these areas.

Various factors control SOC, highlighting the complexity of its dynamics. Understanding the spatial distribution of these variables is crucial for predicting how ecosystems will adapt to changes in climate conditions. Therefore, it is essential to consider the primary predictive drivers across different regions, to effectively model SOC dynamics and ecosystem responses \cite{wang2024upscaling}. Wiesmeier et al. \cite{wiesmeier2019soil} suggested that regional approaches are needed to estimate the SOC storage capacity according to specific climatic conditions and vegetation characteristics. Moreover, another study suggested that SOC dynamics in more than 70\% of topsoil in China were controlled by climate change and its coupled influence with environmental variation \cite{li2022decipher}. In light of these findings, we employed a comprehensive array of input features in our study, including satellite data with a spatial resolution of 30 meters and climate time series data with a spatial resolution of $\sim$4 km. Given that the quality of input features and the uncertainty within the data significantly influence prediction accuracy, further improvements can be achieved by utilizing high-resolution satellite images and/or climate datasets with finer resolutions (e.g. 1 km). This refinement could be particularly crucial for the application of our approach on a regional scale.

\subsection{Future work}

The achieved results demonstrated the better capability of the developed model, SSL-SoilNet, in predicting SOC compared to other implemented ML models. Its applicability extends beyond specific properties and enables accurate estimation of a diverse range of soil characteristics such as soil textures particularly clay, in different geographic regions, ranging from national-scale to larger-scale land management and environmental assessment projects. The flexibility of the proposed model allows other researchers and practitioners to adapt it to their specific soil analysis requirements. This model could also be utilized with hyperspectral data, which offers a large number of spectral bands. This can provide a wealth of surface information features compared to multi-spectral sensors, significantly enhancing the model's prediction ability. Another approach could involve explaining the model's predictions by utilizing summarization methods to combine pattern activations and create attribution graphs. These graphs would highlight the most important spectral features, enhancing the model's interpretability and potentially improving its performance for further analysis. The versatility of our model enables its adaptation to diverse study areas and soil datasets. To facilitate further exploration and adoption of our model, we have made it freely available at: \texttt{https://github.com/moienr/SoilNet}.

\section{Conclusion}

Our research demonstrates an in-depth analysis of self-supervised contrastive learning with a hybrid architecture for predicting SOC. The process involves a two-step training: initially utilizing a vast number of random samples through a hybrid backbone for self-supervised learning, followed by supervised fine-tuning using ground reference samples. Our extensive experiments indicated that this method significantly surpassed traditional supervised learning, especially for skewed soil samples. Moreover, our approach revealed that utilizing a hybrid model, designed to process both image-based and climate-based inputs, effectively extracts pertinent features and captures the spatial patterns and temporal dynamics within the data. Compared to standard ML techniques in soil science, our model showed superior performance. The spatial distribution maps  generated by the developed model offer critical insights into the SOC variability and patterns within the examined areas, aiding in agricultural, land management, and environmental decision-making. Additionally, the versatility and adaptability of the self-supervised model to predict various soil properties and its applicability across different regions highlight its value for future soil analysis.

\section{Acknowledgments}

The authors would like to express their gratitude to Thomas Klein from Max Planck Institute for Intelligent Systems (MPI-IS) and University of Tübingen, Tübingen AI Center, Tübingen, Germany, for his valuable insights that have greatly enhanced the quality of this work. 


\bibliographystyle{IEEEtran}
\bibliography{lib_IEEE}

\end{document}


\begin{minipage}{\textwidth}
    \centering
    \huge
    Supplementary Material for SSL-SoilNet: A Hybrid Transformer-based Framework with Self-Supervised Learning for Large-scale Soil Organic Carbon Prediction
    \vspace{1 em} 
    
    \fontsize{10.5pt}{12pt}\selectfont Nafiseh~Kakhani,
        Moien~Rangzan,
        Ali~Jamali,
        Sara~Attarchi,
        Seyed~Kazem~Alavipanah, 
        Michael~Mommert,
        Nikolaos~Tziolas, and
        Thomas~Scholten\\[4em]
    
    \includegraphics[width=\textwidth]{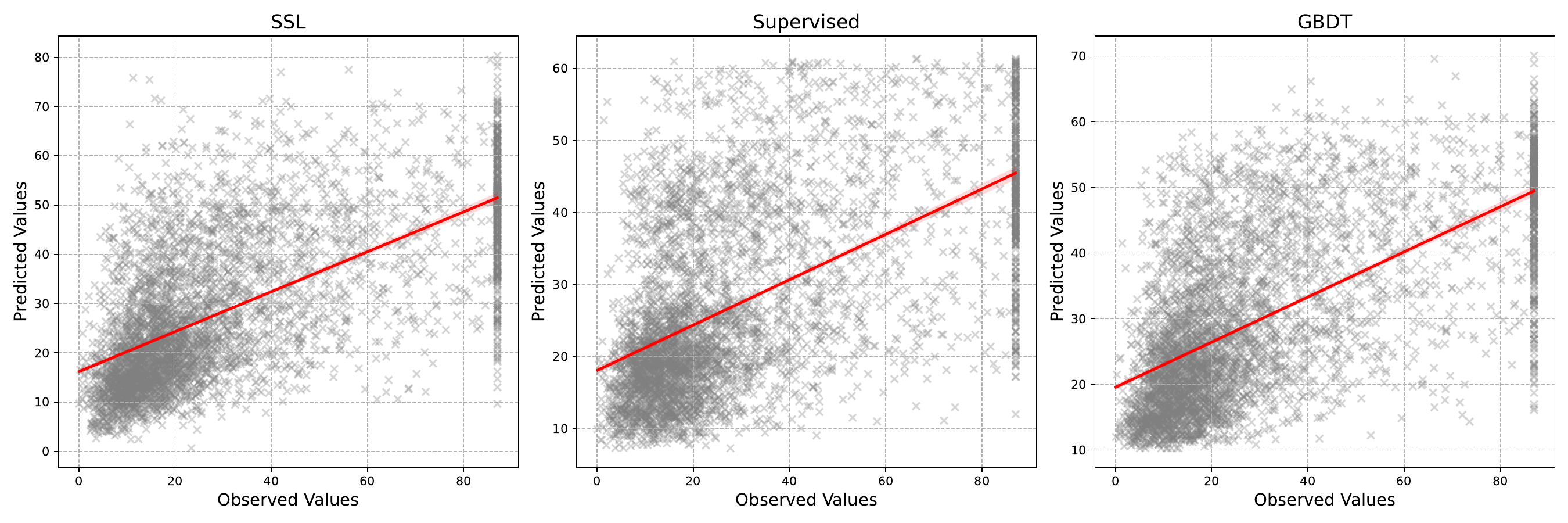}
    \captionof{figure}{Regression plots comparing predicted versus observed values for different implemented models, including SSL and supervised SoilNet, and Gradient Boosting Decision Tree (GBDT) for LUCAS Dataset.}
    \label{fig:S1_LUCAS_reg}
    
    \vspace{5em} 
    
    \includegraphics[width=\textwidth]{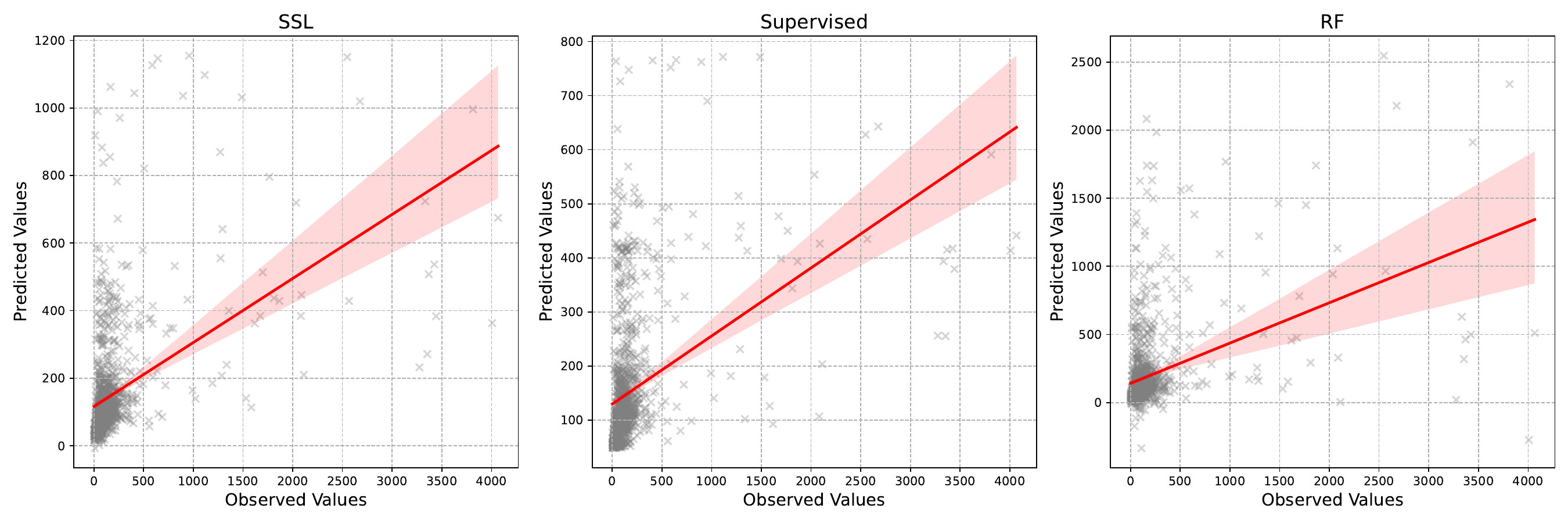}
    \captionof{figure}{Regression plots comparing predicted versus observed values for different implemented models, including SSL and supervised SoilNet, and Random Forest (RF) for RaCA Dataset.}
    \label{fig:S2_RaCA_reg}

\end{minipage}




\begin{figure*}[h]
\centering
\includegraphics[width=7in]{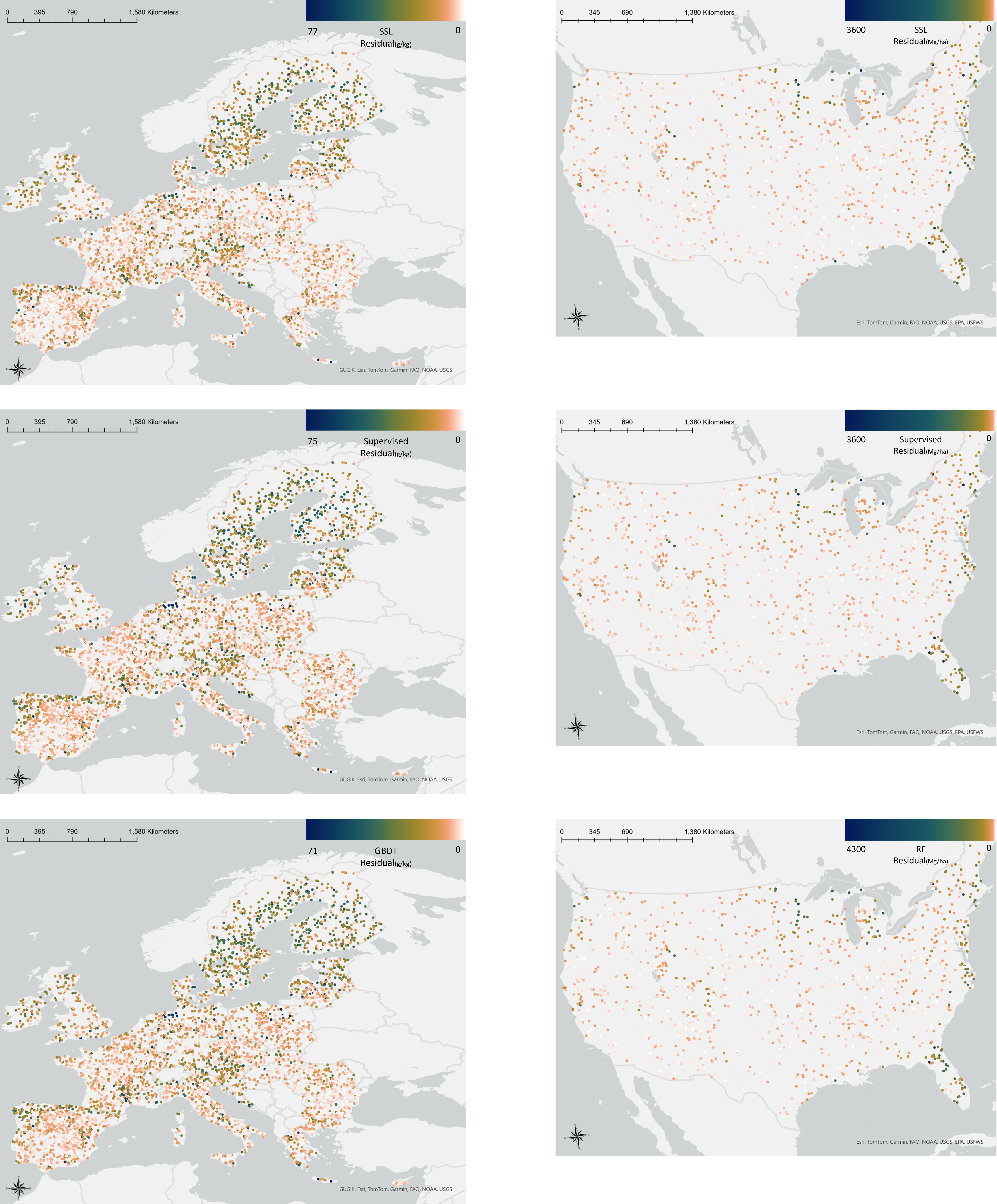}
\caption{Spatial distribution of residuals (absolute error) for Left: LUCAS dataset and Right: RaCA dataset. The first and second rows show the results for SSL and supervised SoilNet, while the last row shows the results for the best ML model (GBDT for LUCAS) and (RF for RaCA).}

\label{fig:error}
\end{figure*}